\documentclass[final,3p,times,twocolumn]{elsarticle}

\usepackage{amsmath,amsfonts,amssymb,latexsym,epic,epsfig,graphics,psfrag,booktabs}
\usepackage{subfigure}
\usepackage{epstopdf}
\usepackage{multirow}
\usepackage{caption}

\journal{Knowledge-Based Systems}

\begin{document}

\begin{frontmatter}



\title{Automated Strabismus Detection for Telemedicine Applications}


\author[a]{Jiewei Lu}
\ead{12jwlu1@stu.edu.cn}

\author[a]{Zhun Fan\corref{cor1}}
\ead{zfan@stu.edu.cn}

\author[b]{Ce Zheng}
\ead{zhengce@hotmail.com}

\author[a]{Jingan Feng}
\ead{13jafeng@stu.edu.cn}

\author[a]{Longtao Huang}
\ead{17lthuang@stu.edu.cn}

\author[a]{Wenji Li}
\ead{liwj@stu.edu.cn}

\author[c]{Erik D. Goodman}
\ead{goodman@egr.msu.edu}

\address[a]{Guangdong Provincial Key Laboratory of Digital Signal and Image Processing, \\College of Engineering, Shantou University,
Shantou 515063, China}
\address[b]{Ophthalmology of Shanghai Children's Hospital, Shang Hai, China}
\address[c]{BEACON Center for the Study of Evolution in Action, Michigan State University, East Lansing, MI 48824 USA}

\cortext[cor1]{Corresponding author}

\begin{abstract}
Strabismus is one of the most influential ophthalmologic diseases in human's life. Timely detection of strabismus contributes to its prognosis and treatment. Telemedicine, which has great potential to alleviate the growing demand of the diagnosis of ophthalmologic diseases, is an effective method to achieve timely strabismus detection. In this paper, a tele strabismus dataset is established by the ophthalmologists. Then an end-to-end framework named as RF-CNN is proposed to achieve automated strabismus detection on the established tele strabismus dataset. RF-CNN first performs eye region segmentation on each individual image, and further classifies the segmented eye regions with deep neural networks. The experimental results on the established tele strabismus dataset demonstrates that the proposed RF-CNN can have a good performance on automated strabismus detection for telemedicine application. Code is made publicly available at: \emph{https://github.com/jieWeiLu/Strabismus-Detection-for-Telemedicine-Application}.

\end{abstract}

\begin{keyword}
Strabismus Detection \sep Telemedicine



\end{keyword}

\end{frontmatter}


\section{Introduction}
Strabismus is an ophthalmologic disease in which eyes can not be aligned at the same location \cite{burian1985burian,van2004amblyopia}, which often occurs in childhood. Strabismus is caused by the problems occurring on optic nerve, brain or extraocular muscle \cite{rutstein2011optometric}. Risk factors contain premature birth and familial inheritance \cite{lorenz2002genetics}. Strabismus has a serious impact on human's life. Strabismus can prevent the brain from merging the two images received from both eyes, which leads to amblyopia \cite{kiorpes1998neuronal}. The under-treated amblyopic eyes may degenerate, resulting in blindness \cite{tommila1981incidence}. Also the double vision and depth perception of strabismus patients are lower than the normal persons. As a result, the prognosis and treatment of strabismus become more and more important in which strabismus detection is the first and one of the most essential steps.

Traditional strabismus detection is performed on the hospital. Doctors use the Hirschberg test \cite{eskridge1988hirschberg} to determine whether the patient has strabismus: a thin beam of light is sent into a patient's eyes for the purpose of verifying whether the reflections of each eye is located at the same place on both corneas. In addition, strabismus detection can be performed with digital tools. In \cite{abrahamsson1986photorefraction}, Abrahamsson et.al. uses photorefraction to achieve small angle strabismus detection. In \cite{loudon2011rapid}, Loudon et.al. utilizes the pediatric vision scanner to perform strabismus detection. In \cite{de2012computational}, Almeida et.al. applies a digital camera and the Hirschberg test to identify strabismus. In \cite{valente2017automatic}, Valente et.al. achieves strabismus detection in digital videos through cover test. In \cite{chen2018strabismus}, Chen et.al. uses an eye tracking system and convolutional neural networks to detect strabismus. These strabismus detection methods have the following disadvantages: (1) It requires the on-site assistance of specialists, and increases the burden of specialists; (2) People in remote districts or isolated communities can not receive timely strabismus diagnosis. In order to overcome the above problems, telemedicine is applied to alleviate the growing demand of strabismus detection. Telemedicine means making use of telecommunication and computer technology to offer clinical health care from a distance. In \cite{helveston2001telemedicine}, Helveston \emph{et.al.} use telemedicine to make the diagnosis of strabismus in the places where specialists are unavailable. In such situation, patients' images were captured with digital cameras, and then sent with computers to specialists so that specialists could make the analysis and diagnosis of strabismus from a distance. However, this method still requires the specialists to spend time in diagnosis.

\begin{figure}
  \centering
  \subfigure[]{
	\includegraphics[width=0.9in,height=1in]{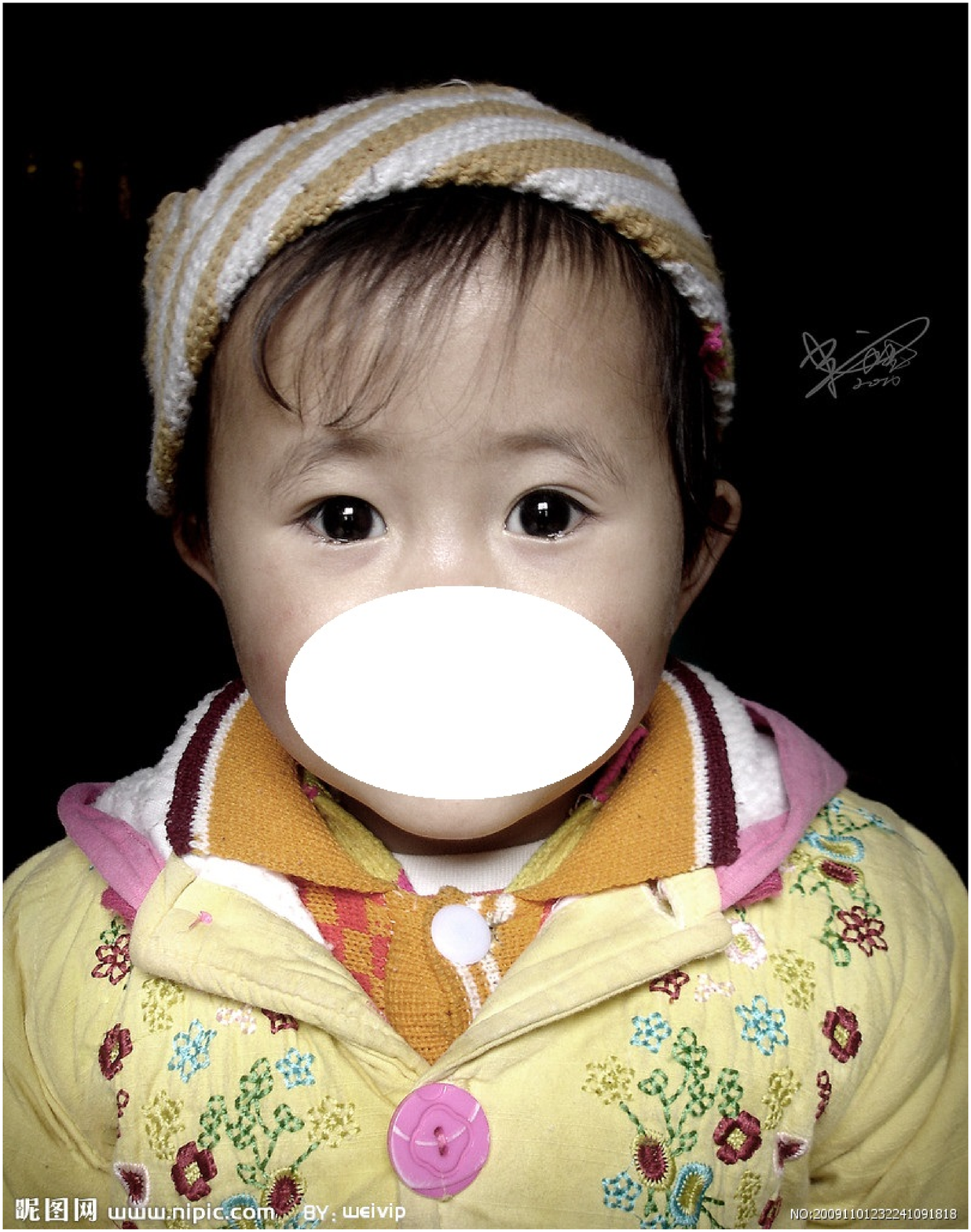}
  }
  \subfigure[]{
	\includegraphics[width=0.9in,height=1in]{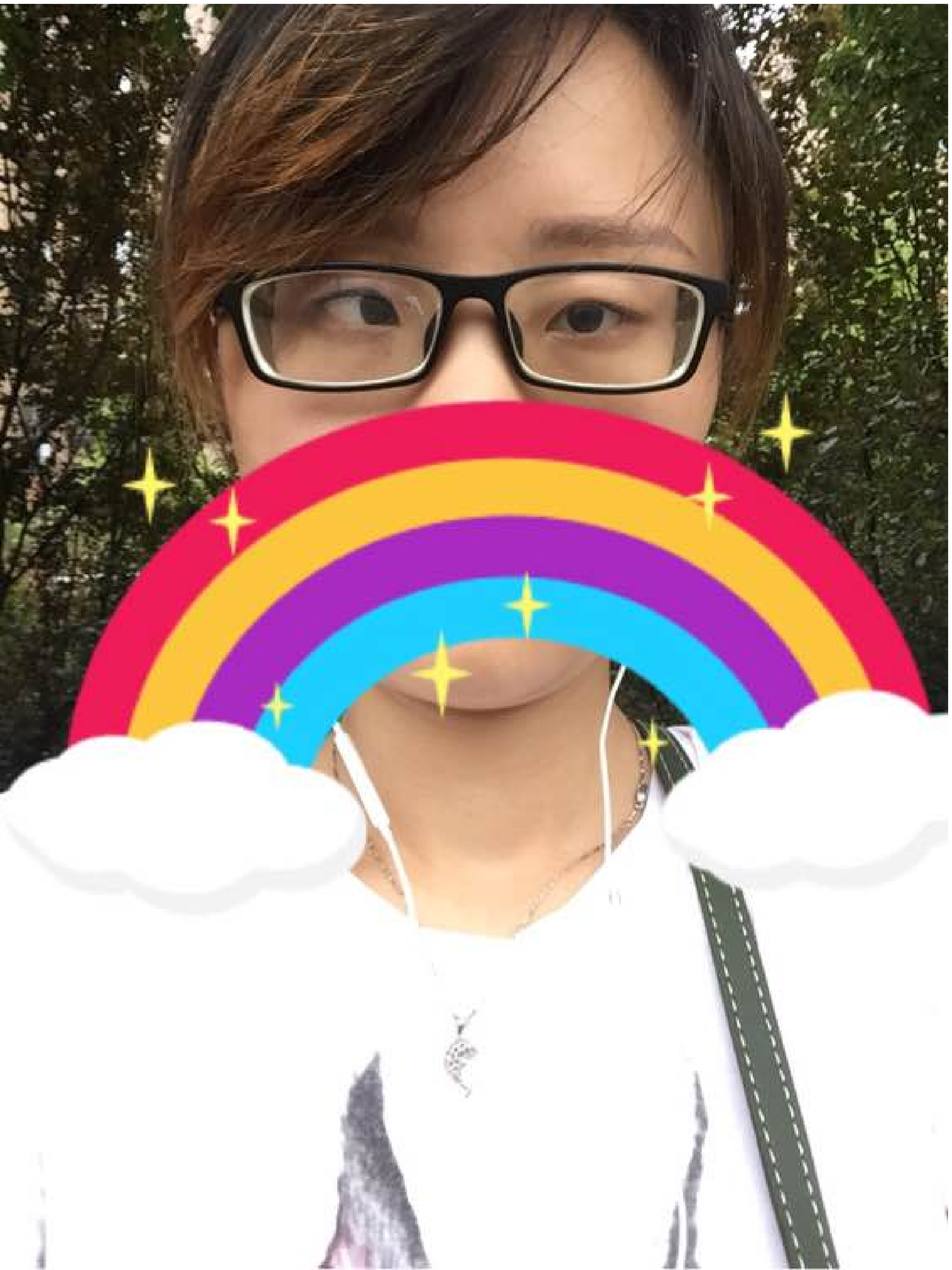}
  }
  \subfigure[]{
	\includegraphics[width=0.9in,height=1in]{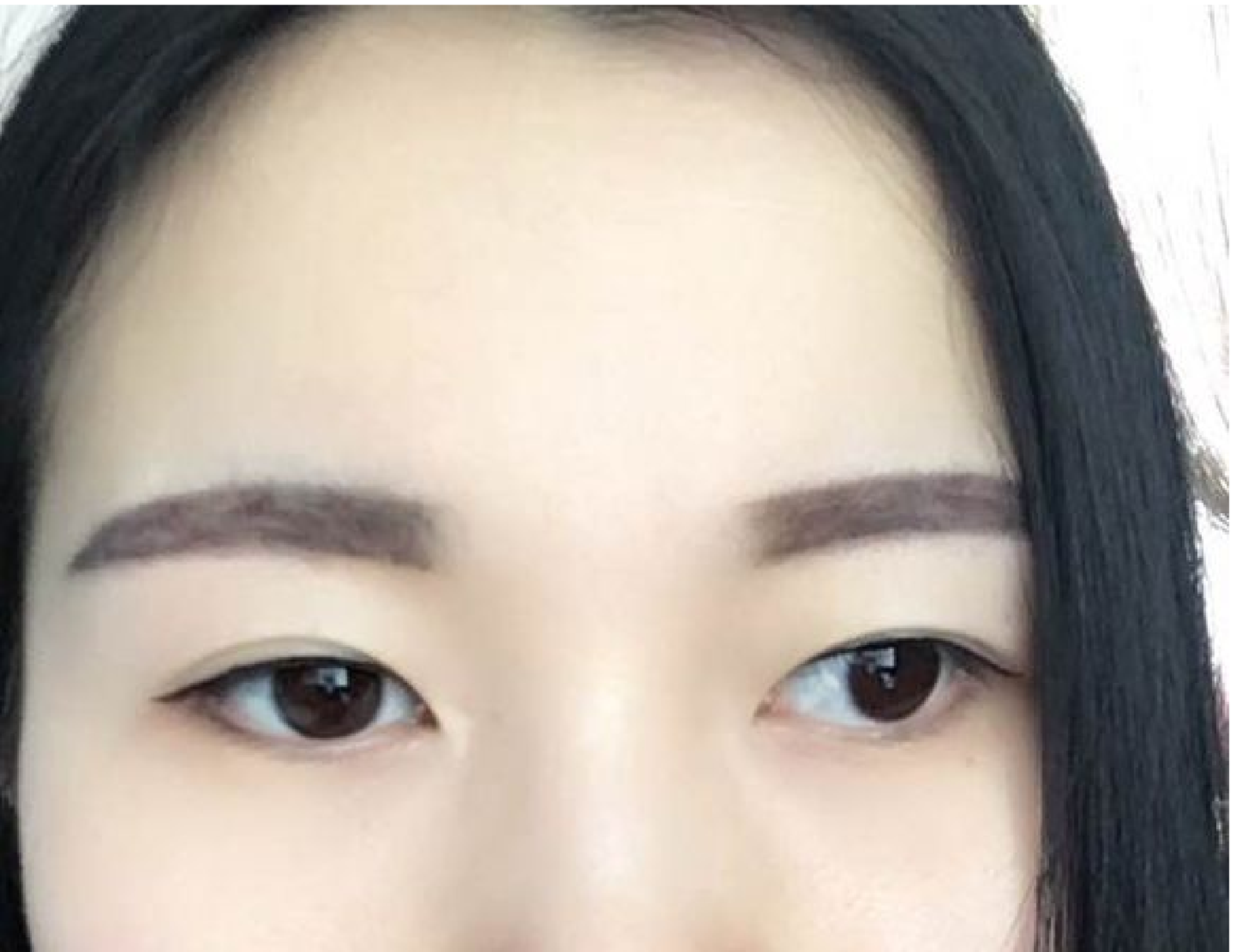}
  }
  \caption{Some exemplary images in the established tele strabismus dataset. (a) The normal image in the dataset. (b) The strabismus image (with a complete human face) in the dataset. (c) The strabismus image (with parts of a human face) in the dataset. }
\label{fig:start}
\end{figure}

To the best of our knowledge, this is the first research of achieving automated strabismus detection for telemedicine application. The major reason is that there are no published tele strabismus datasets. Establishing a tele strabismus datasets is not a trivial task since it needs the collaborations of ophthalmologists and patients. Moreover, the images in the tele strabismus datasets are with different sizes, resolutions and backgrounds, and some images only contain parts of the human faces. These factors make it difficult to achieve automated strabismus detection for telemedicine application.

\begin{figure}
  \centering
	\includegraphics[width=3in]{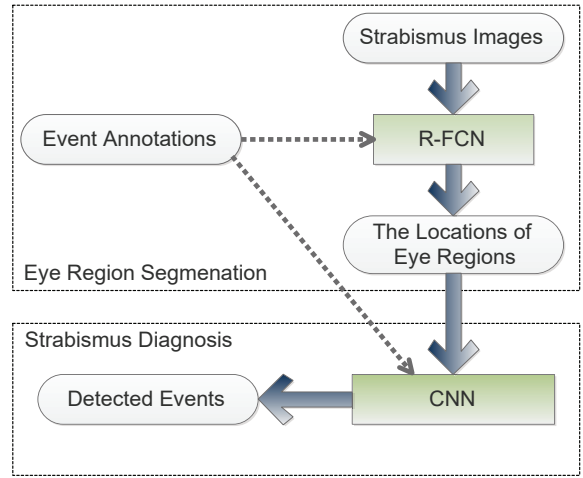}
  \caption{\quad The proposed RF-CNN framework. The dash arrows mean that event annotations are only used in the training stage.}
  \label{fig:whole}
\end{figure}

In order to achieve automated strabismus detection for telemedicine application, in this paper, a tele strabismus dataset is established firstly, which has been carefully collected and labeled by the ophthalmologists. Some exemplary images \footnote{In order to protect the privacies of patients, parts of the human faces are blocked.} in the established strabismus dataset are shown in Figure \ref{fig:start}. Then an end-to-end framework RF-CNN is proposed to perform automated strabismus detection. The proposed RF-CNN comprises two stages: eye region segmentation and strabismus diagnosis (as shown in Figure \ref{fig:whole}). In the eye region segmentation stage, R-FCN \cite{dai2016r} is used to segment the eye regions in the images on the tele strabismus dataset, which aims at reducing the influence of background and focusing on the eye regions. In the strabismus diagnosis stage, automated strabismus detection is achieved according to the segmented eye regions with deep convolutional neural networks (CNN) \cite{lecun1989backpropagation}. The experimental results show that the proposed algorithm can obtain a good detection performance on the tele strabismus dataset.

The rest of this paper is structured as follows. Section 2 introduces some knowledge about telemedicine. Section 3 presents the details of the proposed method.
Section 4 provides the experimental details. In Section 5, the conclusions of this paper are provided.

\begin{figure*}
  \centering
	\includegraphics[width=6in,height=2.5in]{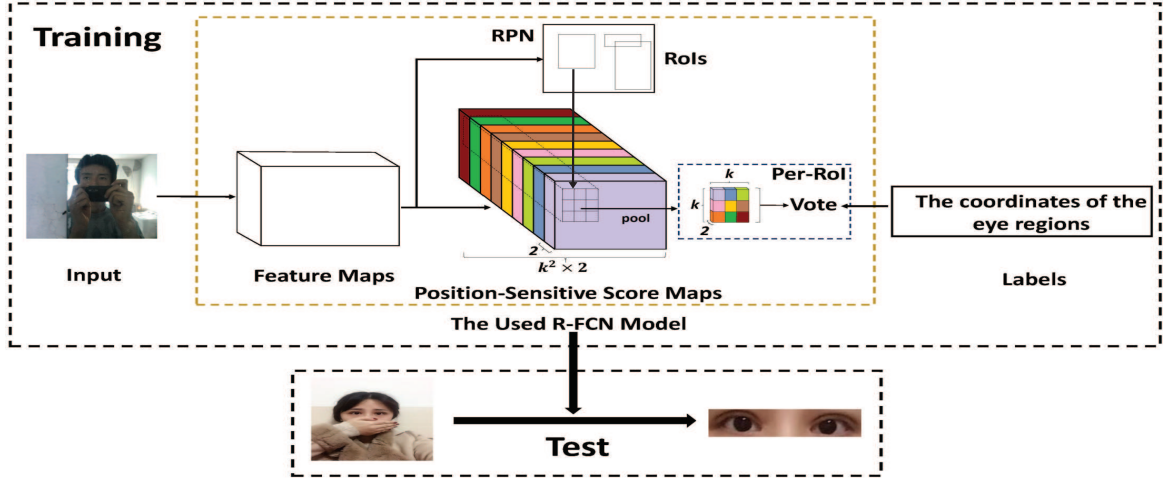}
  \caption{\quad The overall architecture of R-FCN for automated strabismus detection. In our work, the last layer of R-FCN generates ${k\times k=3\times 3}$ position-sensitive score maps. $"2"$ means two categories: eye region and background. Each RoI is divided into $k\times k$ bins, and pooling is only operated on one of the $k\times k$ score maps, which are represented by different colors.}
  \label{fig:R-fcn}
\end{figure*}

\section{Telemedicine}
Telemedicine means providing interactive health care from a distance by using telecommunication and modern technology, which can achieve the transfer of clinical data and provide the remote clinical diagnosis independent of physical proximity to the patient. Telemedicine has an important influence on patients in remote districts and isolated communities since patients can receive health care from doctors (or specialists) far away and do not have to travel to
visit doctors \cite{berman2005technology}. Telemedicine can alleviate the growing demand of diagnosing various diseases, such as diabetic retinopathy and ophthalmologic diseases. It can provide access of specialists to geographically remote districts. Telemedicine consists of three main categories: store-and-forward, remote monitoring and real-time interactive services. Store-and-forward indicates obtaining medical data, and then sending these data to doctors or specialists for offline physical examination \cite{american2012telemedicine}. It is not necessary for doctors and patients to present at the same time. Remote monitoring provides access for specialists to monitor patients remotely with various technological equipments. Real-time interactive services indicates providing real-time interactions between patients and specialists. In this paper, automated strabismus detection belongs to the store-and-forward category.

\section{Methodology}
The proposed framework RF-CNN can be regarded as a two-stage strabismus detection method. As shown in Figure \ref{fig:whole}, the images in the tele strabismus dataset are sequentially fed into R-FCN, which finally outputs the locations of eye regions. Then the eye region images are cropped and fed into CNNs. RF-CNN is directly learned from image data, and the event annotations are used for the training on both stages.

\subsection{Eye Region Segmentation}
\begin{figure}
  \centering
  \subfigure[]{
	\includegraphics[width=0.9in,height=1in]{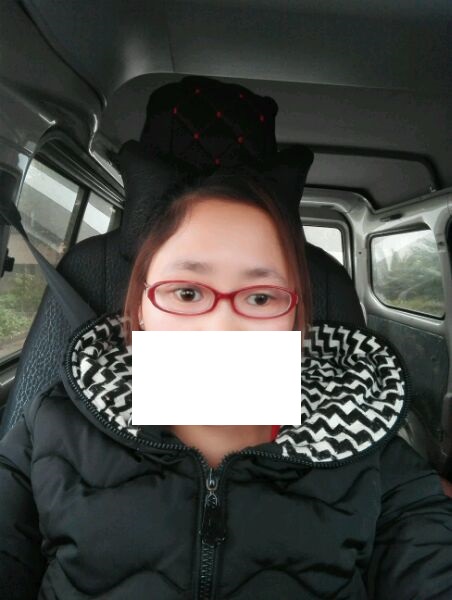}
  }
  \subfigure[]{
	\includegraphics[width=0.9in,height=1in]{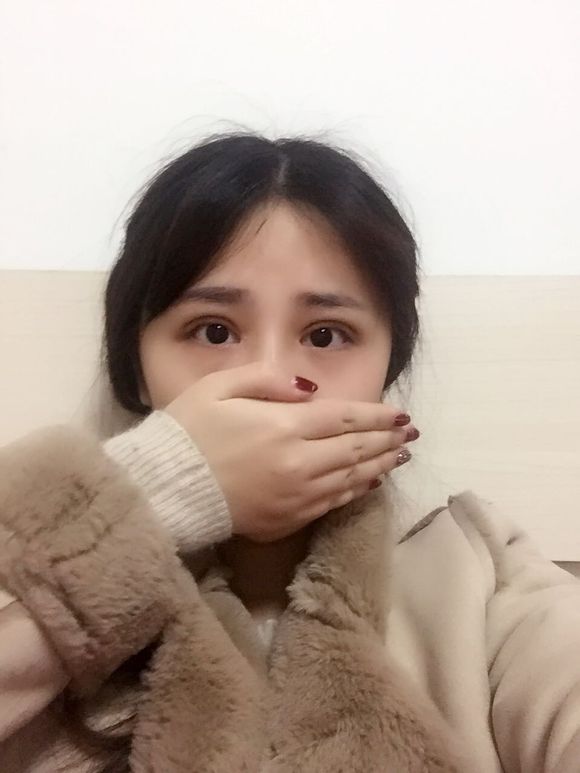}
  }
  \subfigure[]{
	\includegraphics[width=0.9in,height=1in]{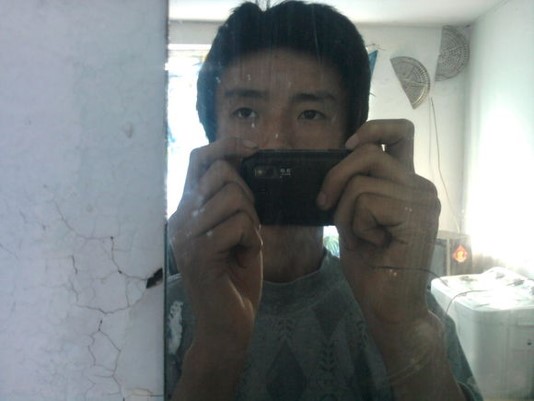}
  }
  \caption{The images with different backgrounds in the tele strabismus dataset: (1) car, (2) wall, (3) home. }
\label{fig:background}
\end{figure}

There are two reasons of performing eye region segmentation: (1) According to ophthalmologists, the diagnosis of strabismus is based on the eye regions of humans; (2) It contributes to reducing the background influence. Figure \ref{fig:background} shows some images with different backgrounds \footnote{In order to protect the privacies of patients, parts of the human faces are blocked.}. R-FCN is built and used to alleviate the above problems in our work. The used R-FCN is composed of two stages: region proposals and region classification. The overall architecture of R-FCN for automated strabismus detection is shown in Figure \ref{fig:R-fcn}. In the region proposals stage, candidate eye regions are obtained by the Region Proposal Network (RPN) \cite{ren2015faster}. After generating the proposal eye regions (RoIs), R-FCN is further dividing the RoIs into the eye region category or the background category. The last layer of R-FCN generates $k^2$ position-sensitive score maps for the eye regions and background, which results in a $k^2\times 2$-channel layer. In our paper, $k$ is set as $3$ \cite{dai2016r} and the 9 score maps encode (top-left, top-center, top-right,..., bottom-center, bottom-right) cases for eye region or background. Then each RoI is voted by averaging the $k^2$ scores:
\begin{equation}\label{equ:scoreAll}
  s_c(P) = \Sigma_{x,y}s_{c}(x,y|P)
\end{equation}
where $s_c(P)$ is the average response for the $c$-th category; $P$ means all learnable parameters in R-FCN; $s_{c}(x,y|P)$ is the response in the $(x,y)$-th bin for the $c$-th category, which is defined as:
\begin{equation}\label{equ:scoreAll1}
  s_{c}(x,y|P) = \sum_{(i,j)\in bin(x,y)} o_{x,y,c}(i+i_0, j+j_0|P)/n
\end{equation}
where $o_{x,y,c}$ is one score map out of the $k^2\times 2$ maps, $(i_0,j_0)$ is the top-left case of an RoI, $n$ means the pixel number in the bin.

Finally, the R-FCN architecture is trained by optimizing the loss function $L$:
\begin{equation}\label{equ:loss}
  L = L_{cls}+L_{reg}^{eye}
\end{equation}
Here $L_{reg}^{eye}$ is the bounding box regression for the eye region as defined in \cite{girshick2015fast}, $L_{cls}$ is the cross-entropy loss for classification:
\begin{equation}\label{equ:loss1}
  L_{cls}=-log(e^{r_{c^*}(P)}/\Sigma_{c^\prime=0}^{1}e^{r_{c^{\prime}}(P)})
\end{equation}
where $c^*$ is the RoI's ground-truth label, $c^\prime=0$ and $c^\prime=1$ represent the background and eye region, respectively.

A well trained R-FCN model is used to perform eye region segmentation, as shown in Figure \ref{fig:R-fcn}.

\begin{figure}
  \centering
	\includegraphics[width=3.3in,height=2.5in]{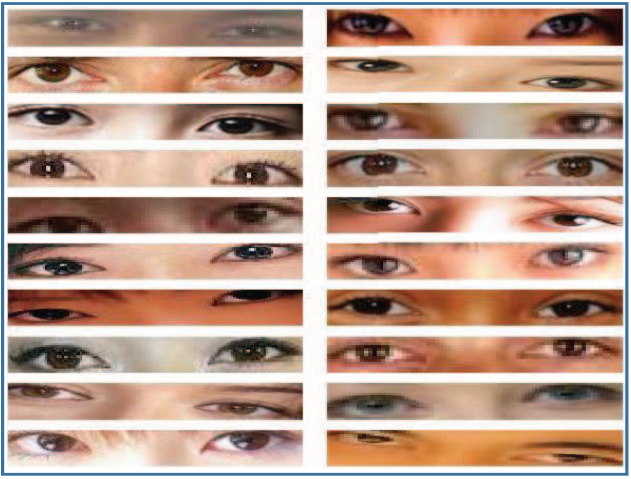}
  \caption{\quad  Some exemplary instances of segmented eye regions with normal labels.}
  \label{fig:normal}
\end{figure}

\begin{figure}
  \centering
	\includegraphics[width=3.3in,height=2.5in]{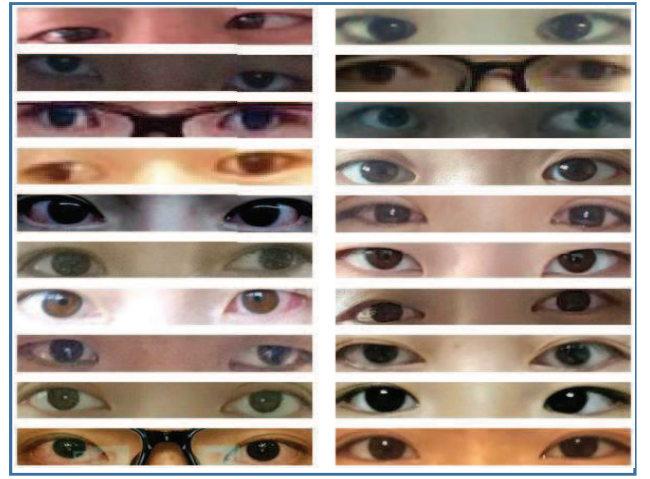}
  \caption{\quad  Some exemplary instances of segmented eye regions with strabismus labels.}
  \label{fig:strabismus}
\end{figure}

\subsection{Strabismus Diagnosis}
Strabismus diagnosis is performed by applying CNNs based on the segmented eye regions. Figure \ref{fig:normal} and \ref{fig:strabismus} show some exemplary instances of segmented eye regions with strabismus and normal labels. CNNs are the commonly used architectures of deep neural networks. They are initially applied to solve the challenging problems like handwritten character recognition \cite{lecun1990handwritten}. Nowadays CNNs have been developed rapidly and used for a a large spectrum of vision problems, such as remote sensing \cite{fan2018automatic} and medical applications \cite{liskowski2016segmenting,zhou2017cell}.

\begin{figure*}
  \centering
	\includegraphics[width=6in,height=1.5in]{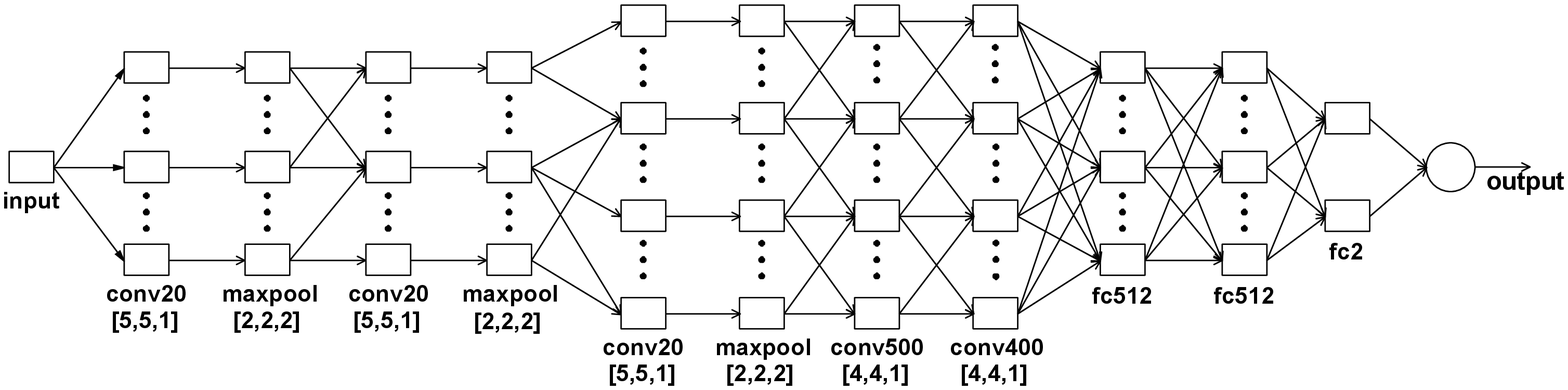}
  \caption{\quad The network architecture composed of five convolutional layers, three pooling layers and three fully connected layers. Layer names are followed by the number of feature maps. Square brackets specify the kernel size and stride. It is noted that 'conv', 'maxpool' and 'fc' are short for convolutional layer, max pooling layer and fully connected layer, respectively.}
  \label{fig:1st}
\end{figure*}

\begin{figure}
  \centering
	\includegraphics[width=3in]{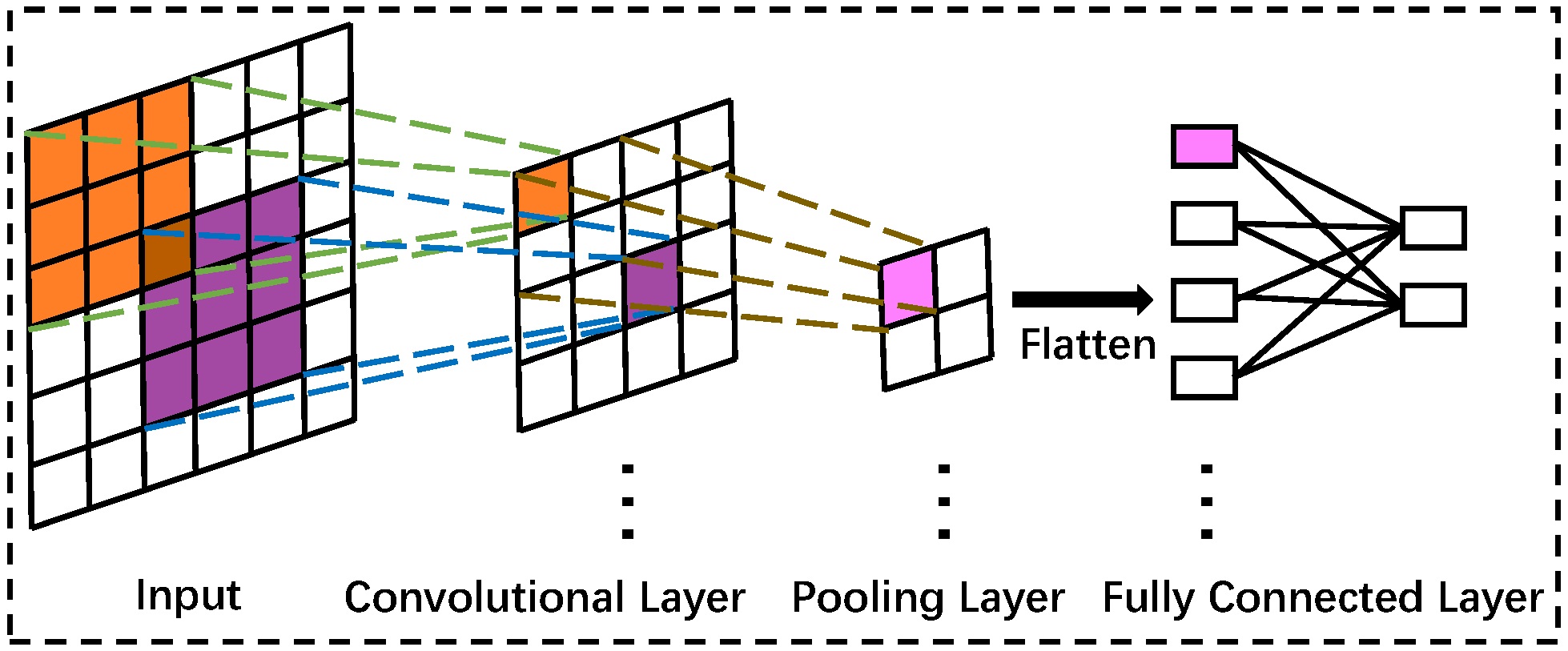}
  \caption{\quad A CNN architecture composed of one convolutional layer, one pooling layer and one fully connected layer. The convolutional layer utilizes a $3\times 3$ kernel with stride $1$, while the pooling layer utilizes a $2\times 2$ kernel with stride $2$. Flatten means converting the feature matrices into vectors.}
  \label{fig:CNNs}
\end{figure}

Generally CNNs consist of convolutional layers, pooling layers and fully connected layers (as shown in Figure \ref{fig:CNNs}). Convolutional layers can extract meaningful and effective features. If the input to the convolutional layer is a $h_i\times w_i\times c_i$ image and the kernel size is $h_k\times w_k$, then the convolutional layer can obtain $k$ features maps with size $[h_i-h_k,w_i-w_k]$. Pooling layer aggregates the neurons' output within a rectangular neighborhood, which can reduce the number of parameters for CNNs. Max-pooling \cite{graham2014fractional} is the most commonly used pooling function. Fully connected layer maps the excitation into output neurons, each corresponding to one decision class.

After confirming the CNN architecture, the parameters $\omega$ of CNNs are learned by optimizing the objective function $J(\omega)$:
\begin{equation}\label{equ:costFunction}
  J(\omega) =  \frac{1}{n} \sum_{i=1}^{n} ||f(\omega,x_i)-y_i||^2_2
\end{equation}
where $n$ is the number of training examples. $(x_i,y_i)$ represents the $i$th training examples. $f(\cdot)$ is the activation function.

\section{Experiments}
The proposed RF-CNN framework is evaluated on an established tele strabismus dataset. In this section, after introducing the tele strabismus dataset, the training and evaluation details of RF-CNN are presented, which includes the training setup, evaluation metrics and experimental results.

\subsection{Datasets}
The tele strabismus dataset contains 5685 images. Each image contains only a human face. 5310 images contain complete human faces while 375 images contain parts of the human faces. The tele strabismus dataset is divided into a training dataset containing 3409 images and a testing dataset containing 2276 images. The training dataset consists of 701 strabismus images and 2708 normal images, while the test dataset is composed of 470 strabismus images and 1806 normal images. These images are captured by various equipments, such as mobile phone and vidicon. In addition, these images have various resolutions, ranging from $1033\times 900$ to $78\times 150$. This dataset has been carefully annotated by the ophthalmologists.

\subsection{Training Setup}
The proposed RF-CNN is trained sequentially. First R-FCN is trained to segment the eye regions. Then a CNN is learned to achieve strabismus diagnosis based on the segmented eye regions. Training setup of RF-CNN is introduced as follows.

R-FCN: ResNet-101 \cite{he2016deep} is adopted as the backone of R-FCN, and online hard example mining (OHEM) \cite{shrivastava2016training} is used to train R-FCN. The learning rate is $0.0003$ and the momentum is $0.9$. A number of eye regions are segmented by R-FCN, which are further resized into $224\times 224\times 3$.

CNN: The network architecture consists of five convolutional layers and three pooling layers, followed by three fully connected layers, as shown in Figure \ref{fig:1st}. Each convolutional layer is followed by a Relu layer \cite{nair2010rectified}, an effective activation function to improve the performance of the CNNs. In addition, the dropout strategy \cite{krizhevsky2012imagenet} is used in the first two fully connected layers in order to prevent overfitting. The network training is performed by the stochastic gradient descent method \cite{bottou2012stochastic}. $L_2$ regularization with the weight decay $5\times 10^{-4}$  is used in the network training. The dropout ratio is set as 0.5. The batch size is set as 32. The learning rate is initially set as 0.01 and the training is stopped after 5000 iterations.

The implementation of RF-CNN was based on Tensorflow \cite{abadi2016tensorflow}, an effective toolbox to train deep neural networks. The training was conducted on a Intel Xeon E5-2690 CPU with a TITAN Xp GPU.

\begin{table*}[htbp]
  \centering
  \caption{The detection performance of RF-CNN on the established tele strabismus dataset}
  \captionsetup{font={small}}
    \begin{tabular}{ccccccccc}
        \toprule
    \toprule
    Metrics & TP    & TN    & FP    & FN        & Se    & Sp    & Acc   & AUC \\
    \midrule
    RF-CNN & 452     & 1685     & 121     & 18     & 0.9330     & 0.9617     & 0.9389    & 0.9865      \\
    \bottomrule
    \bottomrule
    \end{tabular}%
  \label{tab:1st}%
\end{table*}%

\subsection{Evaluation Metrics}
Four commonly used evaluation metrics \emph{Sensitivity}, \emph{Specificity}, \emph{Accuracy} and $\emph{AUC}$ are used in this experiment in order to evaluate the performance of RF-CNN:
\begin{eqnarray*}
      && Sensitivity = \frac{TP}{TP+FN}   \\
      && Specificity = \frac{TN}{TN+FP}   \\
      && Accuracy = \frac{TP+TN}{TP+TN+FP+FN}
\end{eqnarray*}
where \emph{TP} (true positive), \emph{TN} (true negative), \emph{FP} (false positive) and \emph{FN} (false negavtive) are the numbers of correctly identified strabismus images, correctly identified normal images, incorrectly identified strabismus images and incorrectly identified normal images, respectively. Sensitivity (Se) and Specificity (Sp) indicate the RF-CNN's ability of identifying normal images and strabismus images. Accuracy (Acc) is used for evaluating the overall detection performance. In addition, the area under the receiver operating characteristic (ROC) \cite{fawcett2006introduction} curve (AUC) is also applied to evaluate the overall detection performance of RF-CNN.

\begin{figure}
  \centering
\includegraphics[width=3.4in,height=1.7in]{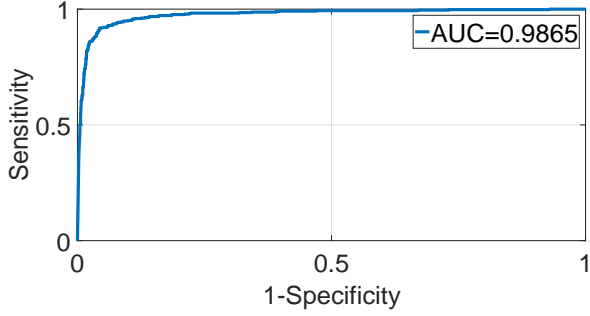}
  \caption{The ROC curve of RF-CNN. }
\label{fig:auc}
\end{figure}

\begin{figure}
  \centering
\includegraphics[width=3.4in]{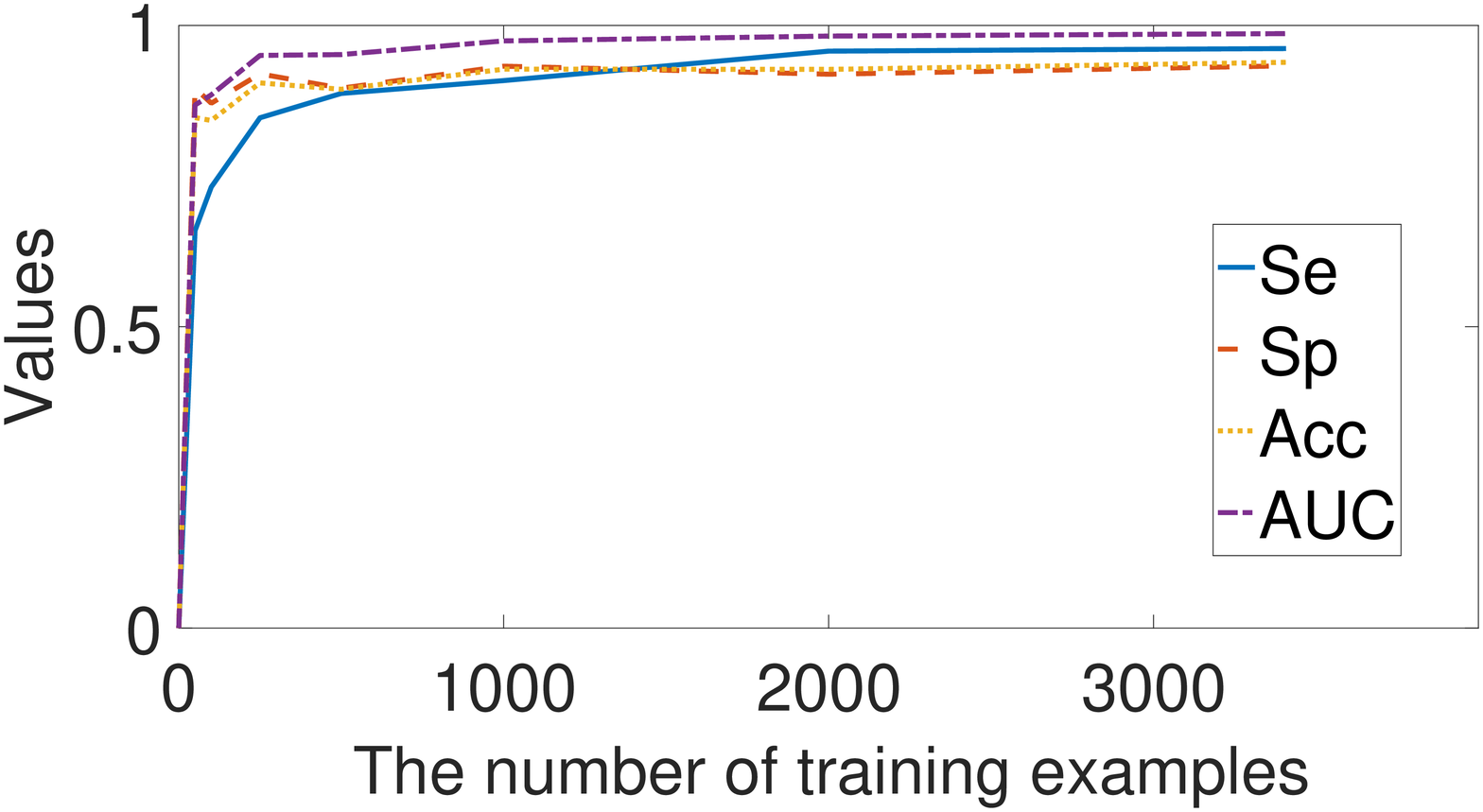}
  \caption{Variations of Se, Sp, Acc and AUC of RF-CNN with the increase of the number of training examples }
\label{fig:3rd}
\end{figure}

\subsection{Results}
The detection results of RF-CNN on the established dataset are shown in Table \ref{tab:1st} and Figure \ref{fig:auc}. It can be observed that RF-CNN can achieve high scores of \emph{Sensitivity=0.9330} and \emph{Specificity=0.9617}, which indicates that RF-CNN can perform well on identifying strabismus images and normal images. The high scores of \emph{Accuracy=0.9389} and \emph{AUC=0.9865} are obtained by RF-CNN, which means that RF-CNN can achieve overall good detection results on the established strabismus dataset.

Moreover, the sensitivity analysis of RF-CNN to the number of training examples is shown in Figure \ref{fig:3rd}. From Figure \ref{fig:3rd}, it can be observed that the evaluation metrics Se, Sp, Acc and AUC become better with the increase of training examples. With less than 1500 training examples, the detection results improve significantly with the increase of training examples. With more than 1500 training examples, the detection results vary slightly with the increase of training examples. From the above observation, we can choose 1500 training examples to train the RF-CNN in our work

\section{Conclusion}
Nowadays strabismus has become an influential ophthalmologic diseases in humans life. Strabismus detection plays an important role in the prognosis and treatment of strabismus. Telemedicine is an effective method to achieve timely detection of strabismus. Concretely, store-and-forward, a category of telemedicine, is applied to achieve timely strabismus detection in this work, which means collecting the medical data, and then sending the data to doctors for the physical diagnosis and examination.

In this paper, in order to achieve automated strabismus detection for telemedicine application, a tele strabismus dataset is established firstly, in which the image data are collected and labeled by the specialists. Then an end-to-end framework RF-CNN is proposed to achieve automated strabismus detection. The proposed RF-CNN first uses R-FCN to perform eye region segmentation, and then classifies the segmented eye regions as strabismus or normal with a deep convolutional neural network. The detection results on the established tele strabismus dataset demonstrate that the proposed RF-CNN performs well on automated strabismus detection for telemedicine application. Moreover, to the best of our knowledge, this is the first research that achieves automated strabismus detection for telemedicine application.

In the future, we will try to release the established tele strabismus dataset by negotiating with doctors and patients. Moreover, we will continue to work closely with doctors, and try to apply this research to clinical applications.

\section{Acknowledgement}
The authors would like to thank Doctor Ce Zheng for providing the tele strabismus dataset.

This research work was supported by Guangdong Key Laboratory of Digital Signal and Image Processing, the National Natural Science Foundation of China under Grant (61175073, 61300159, 61332002, 51375287), the Natural Science Foundation of Jiangsu Province of China under grant SBK2018022017.






%
%
%
\bibliographystyle{elsarticle-num}
\bibliography{StrabisumsDetection}

\end{document}